\documentclass{jdmdh}
\usepackage{array}
\usepackage{pgfplots}
\usepackage{graphicx}
\usepackage{booktabs}
\usepackage{color}
\usepackage{natbib}
\usepackage{graphicx}
\usepackage{titlesec}
\usepackage{float}

\title{Recognition of Oracle Bone Inscriptions by using Two Deep Learning Models}
\author[1]{Yoshiyuki Fujikawa}
\author[1]{Hengyi Li}
\author[1]{Xuebin Yue}
\author[2]{Aravinda C V}
\author[2]{Amar Prabhu G}
\author[1]{Lin Meng}
\affil[1]{Department of Electronic and Computer Engineering,
Ritsumeikan University, Kusatsu,Shiga,Japan} 
\affil[2]{N.M.A.M Institute of Technology, Nitte, India} 

\corrauthor{Lin Meng}{menglin@fc.ritsumei.ac.jp}

\titlespacing*{\section} {0pt}{2.2ex plus 1ex minus .2ex}{2.3ex plus .2ex}

\titlespacing*{\subsection} {0pt}{2.0ex plus 1ex minus .2ex}{1.5ex plus .2ex}

\begin{document}

\maketitle

\abstract{Oracle bone inscriptions (OBIs) contain some of the oldest characters in the world and 
were used in China about 3000 years ago. 
As an ancient form of literature, 
OBIs store a lot of information that can help us understand the world history, 
character evaluations, and more.
However, as OBIs were found only discovered about  120 years ago, 
few studies have described them, and the aging process has made the inscriptions less legible. 
Hence, automatic character detection and recognition has become an important issue. 
This paper aims to design a online OBI recognition system for helping  preservation and organization the cultural heritage. 
We evaluated two deep learning models for OBI recognition, 
and have designed an API that can be accessed online for OBI recognition.
In the first stage, you only look once (YOLO) is applied for detecting and recognizing OBIs. 
However, not all of the OBIs can be detected correctly by YOLO, 
so we next utilize MobileNet to recognize the undetected OBIs by manually cropping the undetected OBI in the image.
MobileNet is used for this second stage of recognition as our evaluation of 
ten state-of-the-art models showed that it is the best network for OBI recognition due to its superior performance in terms of accuracy, loss and time consumption. We installed our system on an application programming interface (API) and opened it for OBI detection and recognition. }


\section{Introduction}

\strut
\vspace{-4ex}

Oracle bone inscriptions (OBIs) were inscribed on cattle bone or turtle shells about 3000 years ago
and contain some of the oldest characters in the world.
The OBIs used in the Shang Period in China were buried in ruins due to 
war in 1046 B.C. and were not discovered until 1899.
Very few studies have described them, and the aging process has made the inscriptions less legible. Understanding the inscriptions is important in terms of researching world history, character evaluation, and more \citep{OBIBOOKOchiai}

\begin{figure}[ht]
\centering
\includegraphics[width=1.0\linewidth]{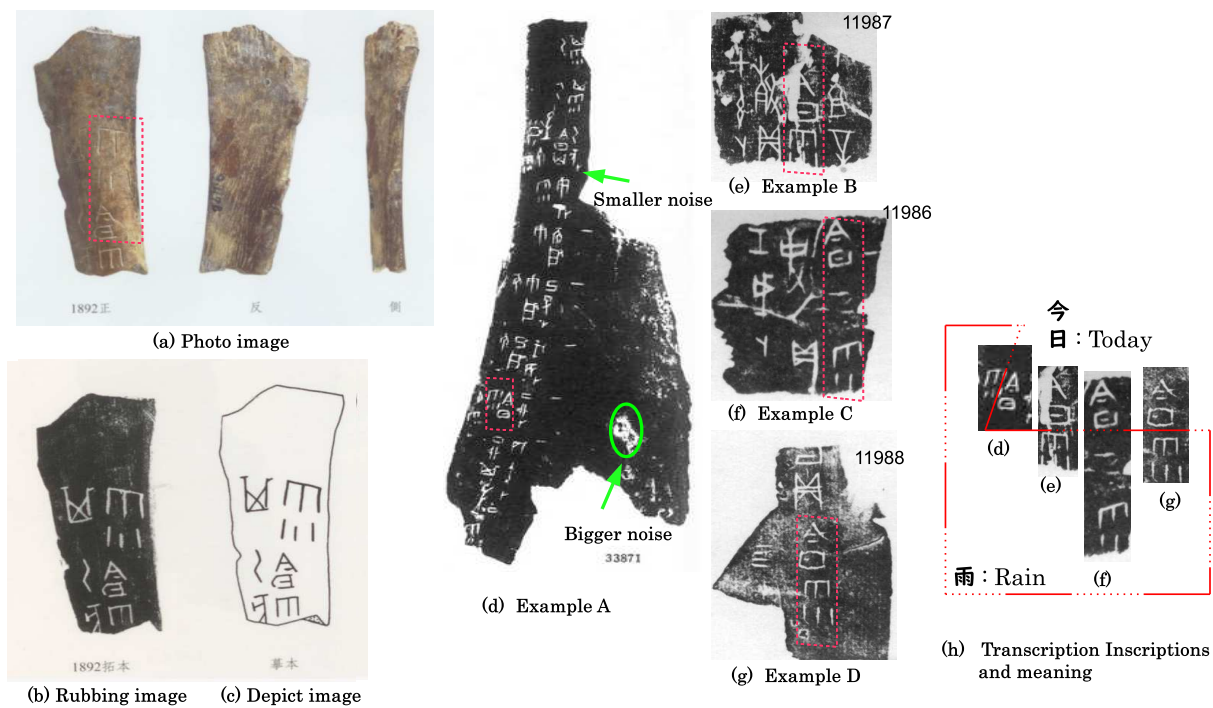}
\caption{Example of OBIs.}
\label{fig:OBI-E}
\end{figure}

The most common OBI retrieval method involves rubbing, where the OBI surface is reproduced by placing a piece of paper over the subject and then rubbing the paper with rolled ink. 
Figure \ref{fig:OBI-E} shows several examples of oracle bones(OBs), and OBIs.
In (a), photo images of the front, side and back of an OB are shown, which several characters visible in the front view image.
(c) and (d) show the rubbing image and an expert depicted image of the front 
side shown in (a), and the characters in the image appear more clearly.
 A search example of OBIs is shown in (d)-(g).
These images were collected by AnYang Normal University and added to OBI Database  \cite{OBI-database}, which is a great resource for culture heritage protection and research.
The phrase ``today 's rain" is indicated by a red box in this example, and the detailed transcription of the inscriptions in modern Chinese along with the meaning in English are shown in (h).

The OBI characters are slightly similar to the modern Chinese characters.
However, this does not necessarily mean that we can understand the OBIs. 
In term of the quality of these characters, the sizes are uneven, both small and big noises exist, 
some characters are broken, and the inclinations of the characters are non-uniform. 
The noises refer the white area in the rubbing image which is not a part of character.
Figure \ref{fig:OBI-E} (d) shows a smaller noise which exists near the character, and a bigger noise in the right side.
These areas like character and interrupt the characters detection and recognition.
   
Hence, OBI detection and recognition are very difficult, and standard optical character recognition (OCR) for OBI recognition is not easy. 
Moreover, character assignments are often different because each of oracle bone is different not only in  size but also in material, 
which leads to difficulty with character detection. 

This paper aims to design a online OBI recognition system for helping  preservation and organization the cultural heritage.
Our proposed method is intended to help archaeologists understand OBI, it features two deep learning models for achieving better OBI detection and recognition.

In the first step, our method applies YOLO \cite{yolov3} for OBIs detection and classification.
However, un-detected OBIs still exist. Next, MobileNet  \cite{MobilNet,MobilNet2} is applied for recognizing the undetected OBIs by manually cropping the undetected OBI in the image.   
We chose MobileNet after the results of our evaluation of ten kinds state-of-the-art models (including   
LeNet \cite{LeNet}, AlexNet \cite{AlexNet}, GoogLeNet \cite{GoogLeNet}, VGG16, VGG19  \cite{VGG}, ResNet 152V2 \cite{Resnet},
Inception \cite{Inception}, InceptionResNetV2 \cite{InceptionResNet}, Xception  \cite{Xception}) showed that it achieved the best performance on loss, accuracy and computation times \cite{MobilNet}.

Finally, the two models are designed as an API that is equip-ped on a server and operated by users.
The API can be accessed in \cite{OBI-API}.

We make three contributions in this paper: 
\begin{enumerate}
 \item This is the first research targeting OBI detection and recognition with real rubbing images using deep learning.
 \item We developed the first online API for OBI recognition, which contributes not only to technology but also to culture heritage preservation and the understanding of OBIs.
 \item Experiments on ten state-of-the-art recognition methods demonstrated that MobileNet is the best model for online OBI recognition.
\end{enumerate}

In section 2 of this paper, we provide an overview of OBIs and discuss related work. Section 3 describes the recognition flow. In section 4, we discuss the first stage of recognition, which detects and recognizes OBIs by YOLO, and present the experimental results. Section 5 describes our selection of the model with the best fit for the second stage of recognition through experiments. Section 6 reports the performance results using our API. We conclude in section 7 with a brief summary and mention of future work.

\begin{figure}[t]
\centering
\includegraphics[width=1.0\linewidth]{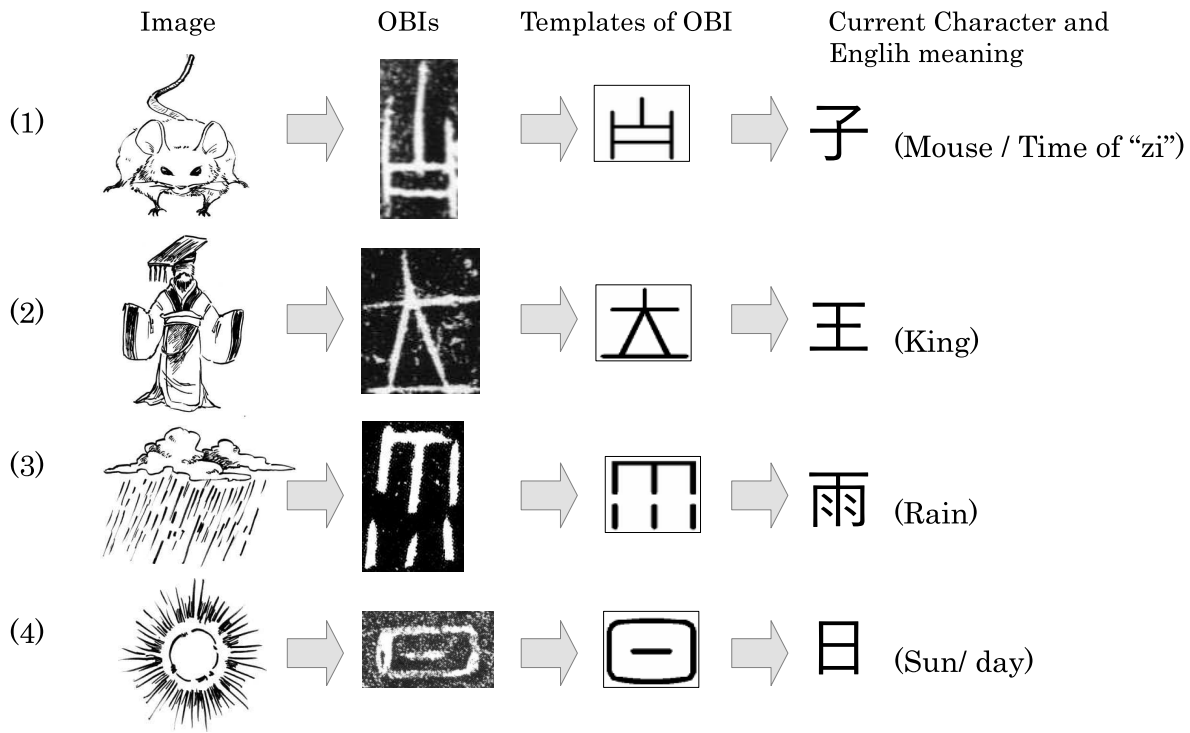}
\caption{Character evaluation.}
\label{fig:mean}
\end{figure}   

\section{OBIs and Related Work}

\subsection{OBIs}
This subsection defines what OBIs are and clarifies the importance of research on their preservation.
OBIs contain some of the oldest characters in the world from the Shang period of China, 
about 3000 years ago. They were inscribed on cattle bone or turtle shells since no ink or paper existed in that period.

OBIs are a kind of hieroglyph that were the original version of the modern Chinese characters, which widely used in China, Japan, and other Asian countries.
One OBI character is one hieroglyph graph, with  a meaning corresponding to one thing or one object.  
Figure \ref{fig:mean} shows several characters and character evaluations, which are introduced in this paper and used in the experiments.
The character evaluation means the changing in writing form during its history of use.
The modern character and the English meaning are shown at the right of the figure.

The first word in Figure \ref{fig:mean} is ``Mouse"(1),  which corresponds  to the image of an animal that has four legs and a long tail. The OBI of ``Mouse" has four legs and a long tail, too. 
The second word is ``King" (2), which corresponds to the image of a man standing on a board (line). 
The OBI of ``King" depicts the same meaning as well. 
The third word is ``Rain", which corresponding image of rain drops falling from a cloud in the sky.
The fourth word is ``Sun(Day)", and the image is a sun.
Understanding these OBIs not only helps us research history but also helps us evaluate the Chinese character evaluations.

To protect these items of cultural heritage, researchers have recently collected OBI images and made them available in books \cite{OBI_rubbing_books} and an open access database \cite{OBI-database}. Ochiai also created a format database for OBIs \cite{OBI_database}.

\subsection{Related Work}

OBI recognition can be defined  as a technique of pattern recognition. 
Pattern recognition methods are typically classified into 
classical computer vision and deep-learning based computer vision.

Various researchers have attempted  to recognize OBIs by means of image processing. 
However, few English papers have reported on OBI recognition, and the recognition rate needs to be improved. 
Li et al. \cite{P_Coding_Principle} presented a recognition method that treats OBIs as a non-directed graph for recording the features of end-points, three-cross-points, five-cross-points, blocks, net-holes, etc. Li et al.\cite{P_Graph_Isomorphism} presented another recognition method using a non-directed graph too.
However, due to the age of OBIs, some of the holes and cross-points that occur are not actually part of the OBIs themselves, which increases the recognition difficulty. Li et al. \cite{P_DNA} proposed a DNA method for recognizing OBIs. However, neither  Li et al. \cite{P_Graph_Isomorphism} nor  Li et al.\cite{P_DNA} provided any details on their experiments.

Meng et al. have previously proposed several methods for recognizing OBIs by template matching and by Hough transform \cite{P_TemplateM,P_HoughM1,P_HoughM2,P_ICIAP}. 
However, the template matching was weak when the original character was tilted, and the tilt was not properly processed in \cite{P_HoughM1}. In one study, there was not enough experimental work \cite{P_HoughM2}. 
Later, Meng et al. proposed a two-stage recognition system to be used from noise reduction to recognition that can consider the tilt of the character; however, only 70\% of the inscriptions on the first-most similar templates were recognized.

Currently, researchers are attempting to recognize the characters by deep learning.
However, work by  Meng et al.  \cite{Eurmed-meng}  and  Liu et al. \cite{OBI_Liu},
showed that all of the OBIs needed to be cut in advance, which is not convenient for practical application. 
In a subsequent work, Meng et al. extended the Single Shot Multibox Detector (SSD) \cite{SSD,OBI-SSD} for OBI detection and 
found through experiments that Precision, Recall and F value achieve 0.95, 0.83 and 0.88 respectively, 
and indicates the effectiveness of SSD.

Another group also aimed to recognize OBIs by deep learning \cite{OBI_Trans}, and while they achieved good accuracy, the experimental images of OBIs were drawn in manual by current researchers, not the original rubbings.

In this paper, we uses two deep learning models for real OBI detection and recognition on real rubbing images.
In the first step, YOLO is applied \cite{yolov3}  for OBI detection and classification.
In the second step, for any OBIs that remain undetected, the user drop a mouse on the image and draws a bounding box around these characters, which are recognized by \cite{MobilNet,MobilNet2}. 
Our models are made openly available to the public by the means of a web-facing API for any users (e.g., archaeologists) who are interested in OBIs. 

\begin{figure}[ht]
\centering
\includegraphics[width=1.0\linewidth]{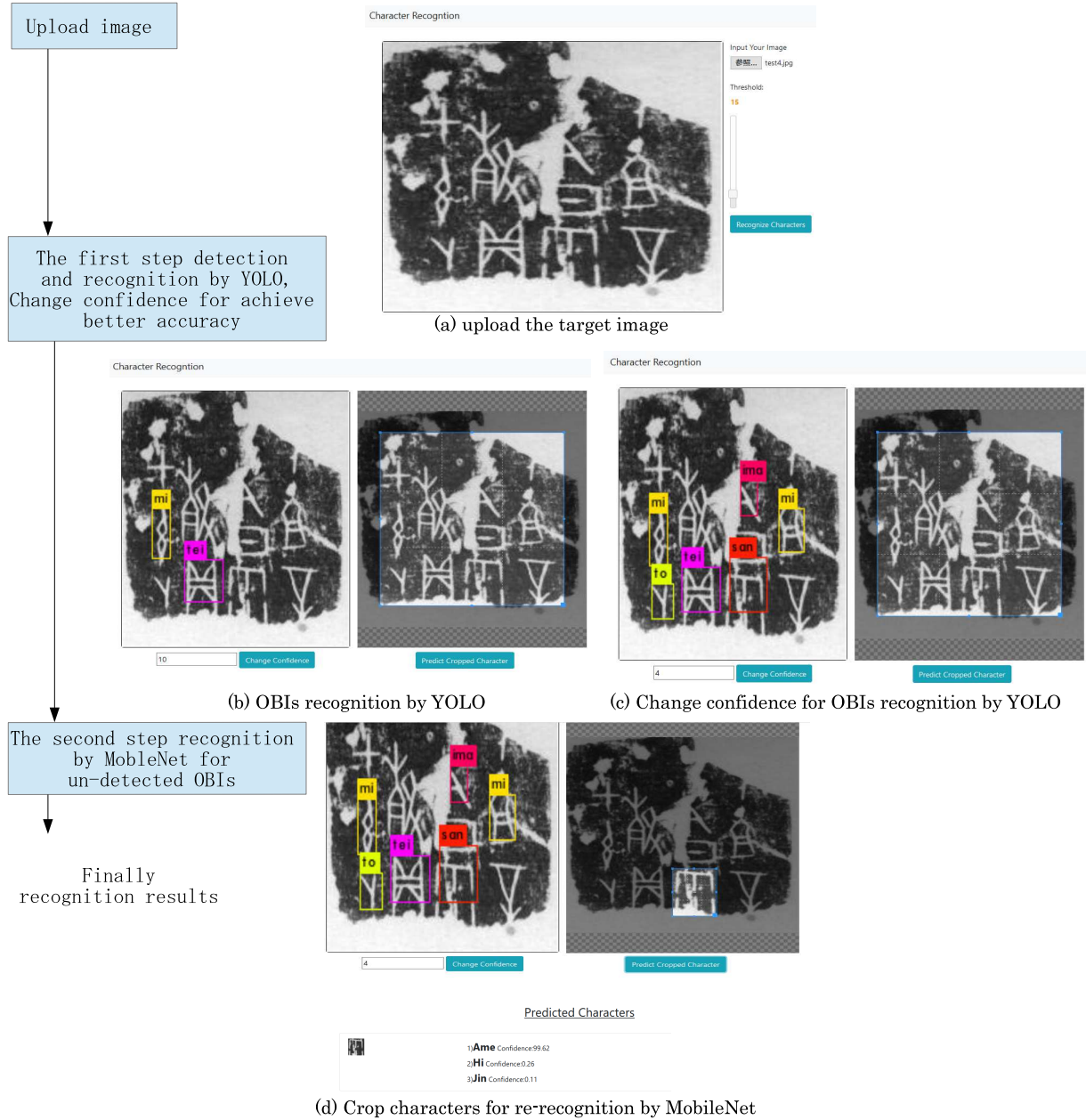}
\caption{OBI recognition flow.}
\label{fig:flow}
\end{figure}

\section{Recognition flow} 
In this section, we provide an overview of the multiple deep learning processes involved in OBI recognition.

The left part of Figure \ref{fig:flow} shows the proposed recognition flow, which consists of two recognition steps after image uploading. 
First, the target OBI rubbing image is uploaded, and OBI detection and recognition by YOLO-tiny is performed using.
YOLOv3 is trained and optimized in advance.
Then, the undetected OBIs are cropped and re-uploaded for re-recognition by MobileNet\cite{MobilNet2,MobilNet}.

These models are designed as an API that is installed on the server for OBI recognition.
The API is opened for accessing and the action is shown in the right part of  Fig. \ref{fig:flow}.
(a) shows the interface for uploading a target OBI rubbing image. Then in(b), when the user clicks the blue ``Recognize characters" button after the target image is uploaded, 
the results recognized by YOLO are shown.
The recognized characters are bounded by boxes and the recognized results and the confidence are displayed on box. 
For achieving better recognition results, the user can change the confidence by clicking the blue ``Change confidence" button. The new result is depicted in Fig. \ref{fig:flow} (c),  which shows the number of recognized characters number increased from  two to six by optimizing the confidence.

 Usually, it is difficult to detect and recognize all of the OBIs by YOLO-tiny,
so we added MobileNet for the character recognition. 
The action, ``Predict Cropped Characters", is shown in (d) . 
The user crops an undetected area (Here, MobileNet is equipped for the characters recognition. The action is shown in  Fig. \ref{fig:flow} (d) of ``Predict Cropped Character". 
 The user crops an undetected area (or any character that needs recognizing) and then clicks the ``Predict Cropped Character" button. The recognized results and the confidence are then displayed on the button of the web site.

\section{YOLO for detecting and recognizing OBIs}
\begin{figure}[t]
\centering
\includegraphics[width=1.0\linewidth]{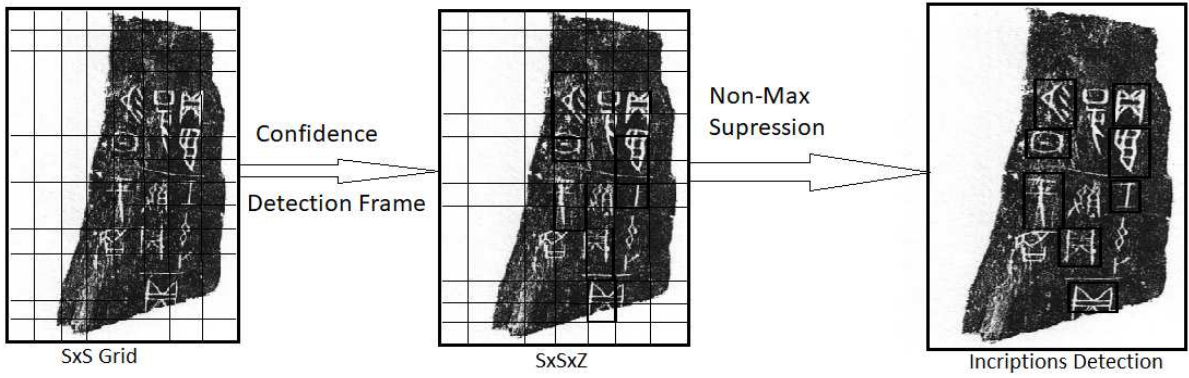}
\caption{Inscriptions detection process of YOLO model.}
\label{fig:YOLO}
\end{figure}

The YOLOv3 tiny network model treats the target detection as a regression problem for a spatially separate target box and its category confidence. A single neural network can directly predict the confidence of the target box and its category from the entire image. The sample detection model of the YOLOv3 network model is shown in the Fig \ref{fig:YOLO}.

We equip a high spec GPU machine for the training, the hardware configuration is GPU : GTX 1050Ti $ \times 3$,   CPU : IntelCore i7 6700 3.4GHz and RAM : 32G; Operating system is Ubuntu16.04 64-bit and  Deep learning framework is  Darknet,  Programming language  is C.

\begin{itemize}
\item The first step is to divide the image into an $ S \times S$ grid. If an inscription is in the grid, the grid is responsible for detecting that inscription. Each grid predicts B detection boxes along with the confidence of these detection boxes. The number of detection frames for each inscription is $ S \times S \times B $.
\item The detection box has five predicted values-- $ X $, $ Y $, $ W $, $ H $, and $ Confidence $ -- where $ X $,$ Y $ is the offset of the center of the prediction box relative to the cell boundary, $ W $ and $ H $ are the ratios of the predicted box width to the entire image, and $Confidence$ represents the confidence of the detection box.
\item Then, each grid predicts the inscription's conditional probability, provided that the known grid contains an inscription.
\item Finally, during detection, the conditional probability is multiplied by the predictive value of different detection box confidences to obtain the score for each box's inscription class. These inscriptions types also include the probability of inscriptions appearing in the detection frame and verify the similarity between the box and the inscription goal. This is formulated by Eq. 1 below.
\end{itemize}
 
\begin{equation}
  Confidence (class) =\mathrm{Probablity}(class) \times  Intersetion \: Over \:     Union_{\mathrm{Predection}}^{Truth}  
\end{equation}

\subsection{Data Set} 
There is no proper and standard OBI data set for training and testing purposes, so we prepared our own. The training and test sets contain 549 and 100 image samples, respectively. We used 27 characters in the experiment. The training data set was used at the initial learning rate of 0.001 and the maximum number of training iterations was about 3000. The mini batch size was ten.
This actually one of the most interesting parts for digital/computational humanties works. More information is necessary here: Show examples of the image samples, what is the distribution of sampled characters, what is the distribution of count of characters per document ``bone", is the dataset a set of bounding boxes, or concrete image samples spliced from the document images.

\subsection{Network Structure Model}
In this work, we use the YOLOv3-tiny network model for training datasets. YOLOv3-tiny follows the coordinate prediction principle of YOLOv2. YOLOv3 can easily achieve real-time performance using just a standard computer with a graphics processing unit (GPU). The Darknet19 structure of the YOLO v3-tiny network is detailed in Table \ref{tab:YOLO} \cite{darknet13}. The given data was augmented to increase its size.

\begin{table}[!ht]
\centering
 \caption{Darknet19 structure of YOLOv3-tiny network.}
  \label{tab:YOLO}
  \begin{tabular}{cccccc}
    Layer & Type & Filters & Size/Stride & Input  & Output  \\ \hline \hline  
    0 & Convolutional & 16 & $3 \times 3$, 1  &  $416 \times 416 \times 3 $ &   $416 \times 416 \times 16$  \\
    1 & Max Pooling   &    & $2 \times 2$, 2  &  $416 \times 416 \times 16$ &   $208 \times 208 \times 16$  \\
    2 & Convolutional & 32 & $3 \times 3$, 1  &  $208 \times 208 \times 16$ &   $208 \times 208 \times 32$  \\
 3 & Max Pooling   &    & $2 \times 2$, 2  &  $208 \times 208 \times 32$ &   $104 \times 104 \times 32$  \\
    4 & Convolutional & 64 & $3 \times 3$, 1  &  $104 \times 104 \times 64$ &   $52 \times 52 \times 64 $  \\
    5 & Max Pooling   &    & $2 \times 2$, 2  &  $104 \times 104 \times 64$ &   $52 \times 52 \times 64 $  \\
    6 & Convolutional & 128 & $3 \times 3$, 1  &  $52 \times 52 \times 64 $ &   $52 \times 52 \times 128 $  \\
    7 & Max Pooling   &     & $2 \times 2$, 2  &  $52 \times 52 \times 128$ &   $26 \times 26 \times 128 $  \\
    8 & Convolutional & 256 & $3 \times 3$, 1  &  $26 \times 26 \times 128$ &   $26 \times 26 \times 256 $  \\
    9 & Max Pooling   &     & $2 \times 2$, 2  &  $26 \times 26 \times 256$ &   $13 \times 13 \times 256 $  \\
   10 & Convolutional & 512 & $3 \times 3$, 1  &  $13 \times 13 \times 256 $ &  $13 \times 13 \times 512 $  \\
   11 & Max Pooling   &     & $2 \times 2$, 1  &  $13 \times 13 \times 512$ &   $13 \times 13 \times 512 $  \\
   12 & Convolutional & 1024 & $3 \times 3$, 1  &  $13 \times 13 \times 512 $  & $13 \times 13 \times 1024 $  \\
   13 & Convolutional & 256  & $1 \times 1$, 1  &  $13 \times 13 \times 1024 $ & $13 \times 13 \times 256 $  \\
   14 & Convolutional & 512  & $3 \times 3$, 1  &  $13 \times 13 \times 256 $ &  $13 \times 13 \times 512 $  \\
   15 & Convolutional & 255  & $1 \times 1$, 1  &  $13 \times 13 \times 512 $ &  $13 \times 13 \times 255 $  \\
   16 & YOLO          &   &   &   &    \\
   17 & Route 13      &   &   &   &    \\
   18 & Convolutional & 128  & $1 \times 1$, 1  &  $13 \times 13 \times 256 $ &  $13 \times 13 \times 128 $  \\
   19 & Up-sampling   &      & $2 \times 2$, 1  &  $13 \times 13 \times 128 $ &  $26 \times 26 \times 128 $  \\
   20 & Route 19 8     &   &   &   &    \\
   21 & Convolutional & 256  & $3 \times 3$, 1  &  $13 \times 13 \times 384 $ &  $13 \times 13 \times 256 $  \\
   22 & Convolutional & 255  & $1 \times 1$, 1  &  $13 \times 13 \times 256 $ &  $13 \times 13 \times 256 $  \\
   23 & YOLO          &   &   &   &    \\ \hline
\end{tabular}
 \end{table}

\subsection{Network Training}
Inscription detectors are trained in accordance with the neural network Darknet framework and the learning rate adopts its default strategy. To reduce the training time, the convolution network is given network parameters that are trained by the Darknet19 network model \cite{darknet13}.

\subsection{Evaluation Metrics}
We used seven metrics for our evaluation: 1) precision, 2) recall, 3) mean of average precision measured at intersection over union (IOU), 4) F1 score, 5) model volume, 6) parameter size, and 7) FLOP. The main confidence and non-maximum suppression threshold for all models in these experiments were set to 0.1 and 0.5, respectively. This evaluation was carried out with no batch processing on one NVIDIA GTX1080ti GPU card using Darknet. Evaluation index metrics were formulated by 

\begin{equation}
    Precision=\frac{TP}{TP+FN}
\end{equation}

\begin{equation}
    Recall=\frac{TP}{TP+FP}
\end{equation}
where TP, FP, and FN respectively represent the number of samples that correctly recognized OBIs as OBIs, the number of samples that identified non-OBIs as OBIs, and the number of samples that identified OBIs as non-OBIs.

\begin{figure}[!ht]
\centering
\includegraphics[width=1.0\linewidth]{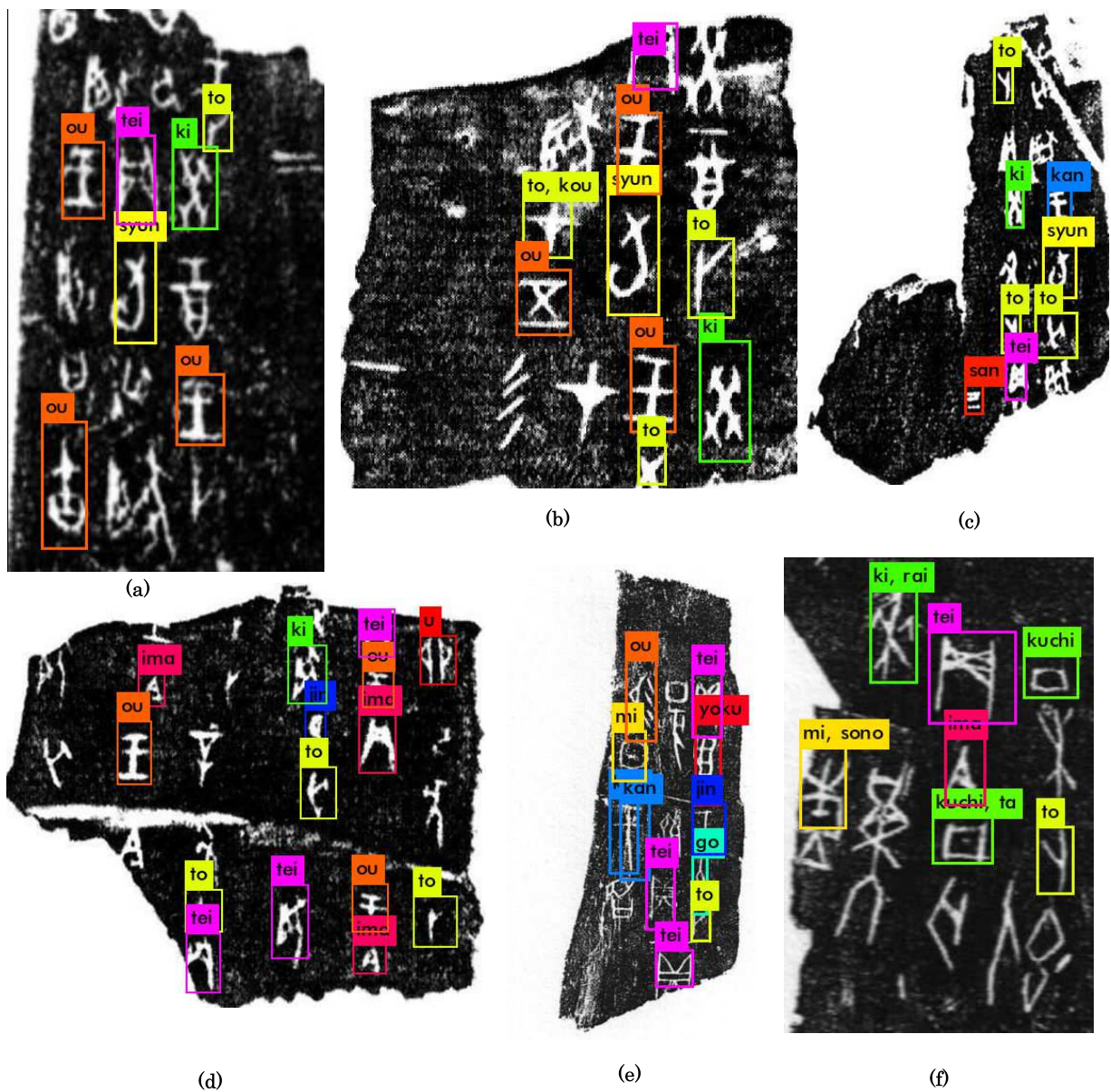}
\caption{Experimental results of YOLO.}
\label{fig:ROBI}
\end{figure}

Six images of the experimental results are shown in Figure \ref{fig:ROBI} as examples. 

\begin{table}[ht]
  \centering
  \caption{Results of two algorithms on OBI test sets.}
  \label{tab:OBITEST}
  \begin{tabular}{ccl}
    \toprule
    Sl.No & SSD & YOLOV3-tiny\\
    \midrule
    Confidence (Eq.1)  & 94.06\% & 90.14\% \\
    Precision (Eq.2)   & 82\%    & 94.06\% \\
    Recall   (Eq.3)    & 88.4\%  & 91.08\% \\
  \bottomrule
\end{tabular}
\end{table}

\begin{figure}[!ht]
\centering
\includegraphics[width=1\linewidth]{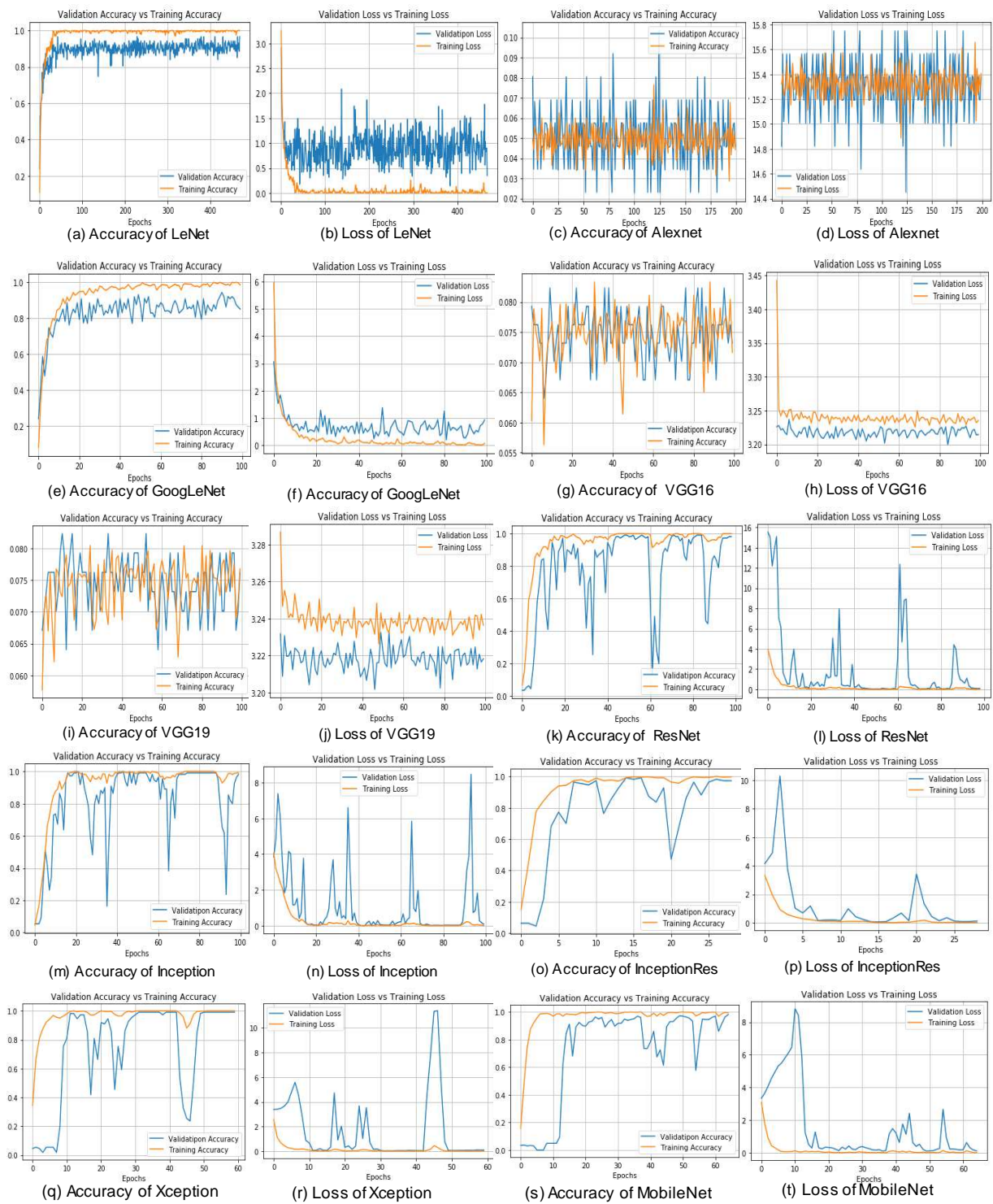}
\caption{Accuracy and loss of ten models.}
\label{fig:AccLoss}
\end{figure}

\begin{table}
\centering
  \caption{Results for OBI classification data}
  \label{tab:COMP-OBI}
  \begin{tabular}{rrrr}
    \toprule
    Model & Epochs & Training accuracy & Validation accuracy \\
    \midrule
     LeNet \cite{LeNet} & 911 & 100\% & 96.30\% \\
     AlexNet \cite{AlexNet} & 200 & 5.21\% & 5.70\% \\
     GoogLeNet \cite{GoogLeNet} & 94 & 98.96\% & 92.05\%\\
     VGG19  \cite{VGG} & 42 &7.52\% & 3.41\% \\
     VGG16 \cite{VGG}  & 15 &6.94\% & 7.55\% \\
     ResNet152V2 \cite{Resnet}& 40 & 96.88\% & 96.59\% \\
     Inception \cite{Inception}& 31 & 95.26\% & 96.59\% \\
     InceptionResnetV2 \cite{InceptionResNet}& 25 & 96.09\% & 97.73\% \\
     XCeption  \cite{Xception}& 52 & 98.18\% & 97.73\% \\
     MOBILE-NET \cite{MobilNet}& 95 & 99.30\% & 98.89\% \\
    \bottomrule
\end{tabular}
\end{table}

\subsection{Experimental results of Loss and Accuracy}
We used 29 classes OBI for the model selection experiment.
The total number of training images was 605.
The dataset are created and stored in \cite{OBI_data}.  
Since OBI are ancient characters, some characters appeared with a low frequency. As such, the data of every class is imbalanced, and the image number of every class varies from two to 46 pieces.

Figures 6 -- 15 show the accuracy and loss of the ten models we evaluated (LeNet, AlexNet, GoogLeNet, VGG16, VGG19, ResNet152V2, Inception, InceptionResNetV2, Xception, and MobileNet, respectively), and the results are summarized in Table \ref{tab:COMP-OBI}. 

In each figure, the left side shows the accuracy of training and validation and the right side shows the loss of training and validation. The horizontal axis shows the number of epochs and the vertical axis shows the accuracy or loss. 

The loss and accuracy of training and validation did not converge in the case of AlexNet, VGG16, and VGG19. This was because the data set was too small to fit the models.

LeNet and GoogLeNet, which are a relatively old models, achieved a better performance in training. However, the validation accuracy was not impressive, especially for GoogLeNet. This means that these models cannot achieve a better performance, but only increase the layer numbers.

In terms of ResNet and Inception, the accuracy and loss of training and validation had better results. Since these newer models -- specifically, InceptionResNetV2, Xception, and MobileNet -- achieved similarly impressive results compared to the other models, it was difficult to select the best model from among the three.


\subsection{Experimental Results of Recognition Time}
In this experiment, we installed the InceptionResNetV2, Xc-eption, and MobileNet models on the API server to measure the recognition time for making the model decision. The CPU is Intel(R) Xeon(R) CPU E5-1410 v2 @ 2.80 GHz, RAM is 8G, and OS is Ubuntu 18.04.3 LTS. The Apache HTTP server was used for designing the API website. 

Table \ref{tab:Time} shows the computation times of the three models. We found that MobileNet consumed the shortest time, which was three times faster than Xception and almost five times faster than Inception ResNetv2. These results demonstrate that MobileNet is the best fit for an embedded system that has a short time consumption.

As MobileNet achieved the best performance in terms of accuracy, loss, and time consumption, this is the model we selected for the second stage of recognition in the API.

\begin{table}
\centering
  \caption{Recognition time on server machine}
  \label{tab:Time}
  \begin{tabular}{cc}
    \toprule
    Model & Recognition time (seconds)\\
    \midrule
     Xception           & 30.05 \\
     InceptionResNet v2 & 43.15 \\
     MobileNet          & 9.55 \\
  \bottomrule
\end{tabular}
\end{table}

\section{Experimentation Results and Discussion}
Figure \ref{fig:flow} shows examples of experimental results by the two deep learning models output by the API. The left one shows the results by YOLO. Several characters were detected and recognized correctly. However, a few characters were not detected, e.g., the first and last characters of the second line in (A). The reason is that part of the first character is broken and the last characters are too large. Also, the left characters in (A) were recognized as two different characters with different sizes. The same problem occurred when MobileNet was used, as shown in the example in (B). However, the undetected characters were recognized correctly by MobileNet after cropping the characters in the rubbing image.

In terms of the recognition of undetected characters, currently, the small size of our dataset limited the performance of YOLO and made it impossible for MobileNet to recognize all of the OBIs. Even so, the effectiveness of the proposed method has been demonstrated. Our future work will involve increasing the size of the dataset.


\section{Conclusion}
Oracle bone inscriptions (OBIs), which were created around 3000 years ago, contain some of the oldest characters in the world. Many of these OBIs are difficult to recognize today due to their age, and understanding them is important in terms of researching world history and evaluating characters. This paper proposed two deep learning models for OBI recognition and designed an online API on which they can be deployed. For the first stage of recognition, YOLO is applied for detecting and recognizing OBIs. Due to the limitations of YOLO, not all of the characters can be detected and recognized correctly. For the recognition of the undetected OBIs, we evaluated ten different models to determine which one is the best fit. Results showed that MobileNet had the best performance in terms of accuracy, loss, and short time consumption. Hence, we also equip the API with MobileNet for cropping any undetected OBIs and re-recognizing them again (the second stage of recognition). At present, our dataset is too small, which limited the effectiveness of the proposed method. Increasing its size will be the focus of future work.

\section*{ACKNOWLEDGEMENTS}
This work was supported by a Grant-in-Aid for Scientists (18K18337) from the Japan Society for the Promotion of Science (JSPS), the Ritsumeikan University Art Research Center and Key Laboratory for Oracle Bone Script Information Processing (Anyang Normal University).

\newpage

\bibliographystyle{plainnat}
\bibliography{jdmdh-example}

\end{document}